%% file: neurips_2022_upload.tex
\documentclass{article}

\usepackage[preprint, nonatbib]{neurips_2022}

\usepackage{bbm}
\usepackage{graphicx}
\usepackage{amsmath}
\usepackage{enumitem}

\usepackage[utf8]{inputenc} 
\usepackage[T1]{fontenc}    
\usepackage[ hypertexnames=false, bookmarksnumbered = True, colorlinks=true, allcolors=blue]{hyperref} 

\usepackage{url}            
\usepackage{booktabs}       
\usepackage{amsfonts}       
\usepackage{nicefrac}       
\usepackage{microtype}      
\usepackage{xcolor}         
\usepackage[OT1]{fontenc}
\usepackage{amsmath,amssymb,amsfonts}
\usepackage{algorithmic}
\usepackage{graphicx}
\usepackage{textcomp}
\usepackage{multirow}
\usepackage{threeparttable}
\usepackage{subfigure}
\usepackage[T1]{fontenc}    
\usepackage{hyperref}       
\usepackage{url}            
\usepackage{booktabs}       
\usepackage{amsfonts}       
\usepackage{nicefrac}       
\usepackage{microtype}      
\usepackage{xcolor}         

\usepackage{cleveref} 
\crefname{assumption}{Assumption}{Assumptions}
\crefname{definition}{Definition}{Definitions}
\crefname{lemma}{Lemma}{Lemmas}
\crefname{theorem}{Theorem}{Theorems}
\crefname{corollary}{Corollary}{Corollaries}
\crefname{proposition}{Proposition}{Propositions}
\crefname{claim}{Claim}{Claims}
\crefname{subclaim}{Subclaim}{Subclaims}
\crefname{procedure}{Procedure}{Procedures}
\crefname{algorithm}{Algorithm}{Algorithms}
\crefname{example}{Example}{Examples}
\crefname{figure}{Figure}{Figures}
\crefname{section}{Section}{Sections}
\crefname{appendix}{Appendix}{Appendices}
\crefname{table}{Table}{Tables}
\crefname{equation}{}{}
\crefname{appsec}{Appendix}{Appendices}
\crefname{fact}{Fact}{Facts}

\title{QML for Argoverse 2 Motion Forecasting Challenge}

\author{
  Tong Su \quad \quad Xishun Wang \quad \quad Xiaodong Yang\\
  QCraft
}

\usepackage{tikz}

\begin{document}

\maketitle

\begin{abstract}
To safely navigate in various complex traffic scenarios, autonomous driving systems are generally equipped with a motion forecasting module to provide vital information for the downstream planning module. For the real-world onboard applications, both accuracy and latency of a motion forecasting model are essential. In this report, we present an effective and efficient solution, which ranks the 3rd place~\cite{wilson2021argoverse} in the Argoverse 2 Motion Forecasting Challenge 2022.
\end{abstract}

\section{Introduction}

As a core component of an autonomous vehicle (AV), motion forecasting or trajectory prediction plays a crucial role to understand the behaviors of traffic agents. A prediction module leverages the perception information~\cite{luo2021simtrack} 
and outputs multi-modal future trajectories for nearby agents. However, this is a challenging
task due to the uncertainty of traffic actors and complexity of road topology~\cite{varadarajan2021multipath++}. 
In this report, we present an effective and efficient approach, 
which forecasts the future trajectories of multi-class agents around the ego vehicle. 
Specifically, our model consists of five parts: (1) agent history encoder that takes the 
agent-centric data as input, (2) agent interaction encoder dynamically models the interactions of 
traffic agents, (3) vector map encoder making use of the local vector map to provide 
the scene context~\cite{liang2020learning}, (4) anchor decoder and (5) 
prediction decoder which receive the context features and respectively produce the proposals and future predictions.

\section{Methodology \label{sec3}}

\begin{figure*}[htbp]
        \centering
        \includegraphics[width=\textwidth, height=7cm]{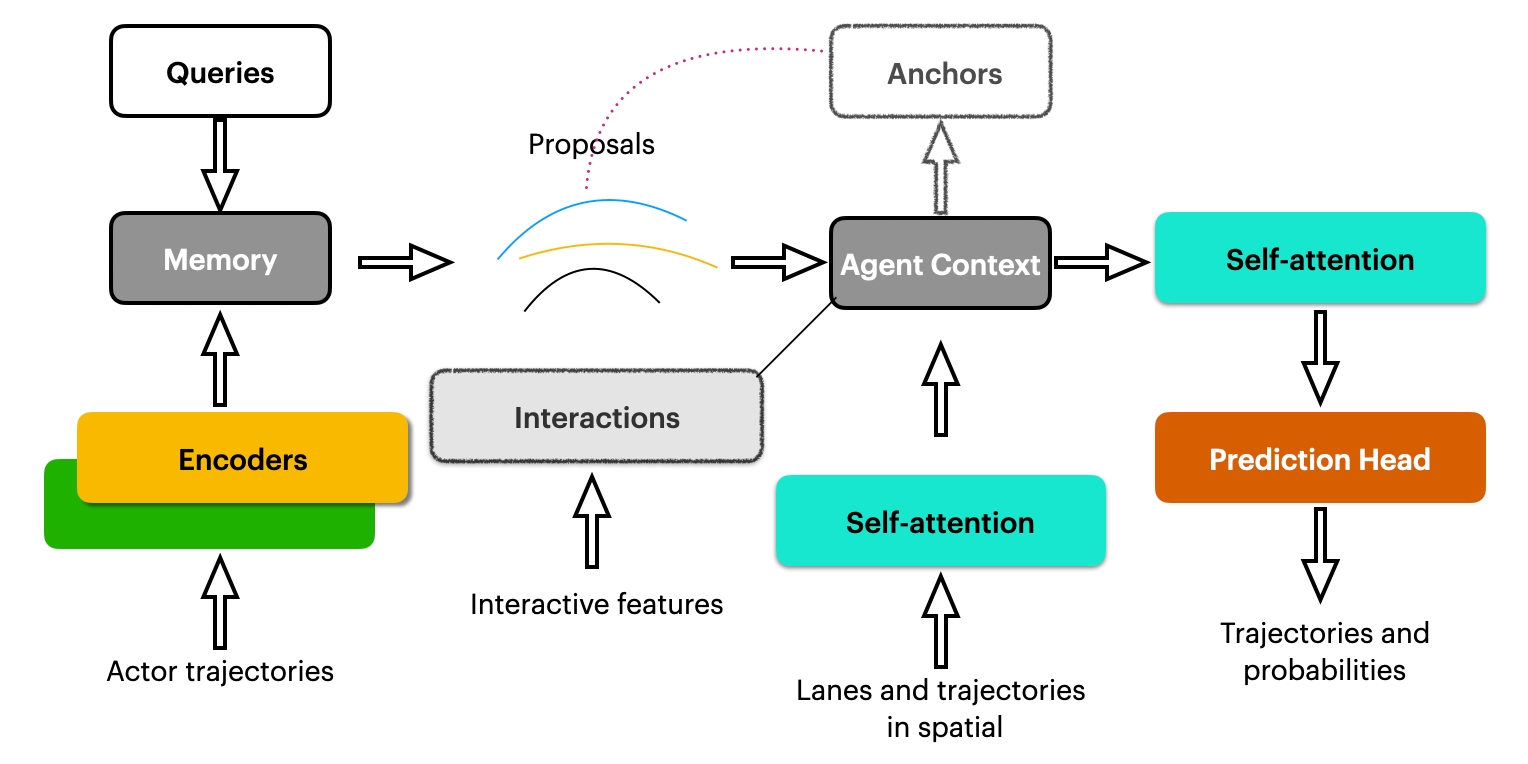}
        \caption{Illustration of the proposed approach. Encoders extract the temporal features and interaction features of history observations. 
        We fuse the dynamic features to get the proposals, and then use self-attention to encode both trajectories and lanes to obtain the 
        spatial context features. Together with the proposals, the agent context that is from the spatial context features are decoded to anchors. Finally, 
        the prediction head outputs the multiple predicted trajectories and probabilities.}
        \label{fig1}
        \end{figure*}

A trajectory prediction model aims to predict the trajectories of agents in the future $T$ timesteps based on the history observation of their past $H$ timesteps. In this section, we will describe our proposed approach as illustrated in Figure~\ref{fig1}.

\subsection{Agent History Encoder}

In order to accurately model the history observation of traffic agents, we first apply an agent-centric paradigm that coverts the trajectory of each agent to an ego-view coordinate system. The history observation contains trajectory, velocity, heading and object type. 
By taking advantage of this agent-specific ego-view, we standardise each actor trajectory such that 
the last timestep of history observation is always at the origin of coordinate system.

\subsection{Agent Interaction Encoder}

In the complex traffic scenarios, agents plan their future routes by interacting with their neighbors~\cite{lee2017desire}. Therefore, modeling the interaction between agents is beneficial to 
predict their future trajectories. The agent interaction encoder takes the interaction 
information as input, which includes the relative position, velocity and heading. 
It is noted that we also convert the interaction input to the agent-centric coordinate system.

\subsection{Vector Map Encoder} 

While the history observations of different agents are informative, the high-definition (HD) map further provides the complementary social context for agents to better plan their future trajectories. For the vector map, we pre-process each single vector map to lane segments. 
By splitting the vector map we can improve our vector map encoder both in accuracy and efficiency. For a single agent, we choose the lane segments within $D$ meters. 
Note that $D$ varies for different agent categories. Furthermore, we transform the lane segments to 
each agent view and construct a spatial topology feature. After that, the lane segment features 
are encoded into the social context~\cite{ngiam2021scene}.

\subsection{Anchor Decoder and Prediction Decoder}

Given the aforementioned features, we adopt the transformer network to integrate the social context~\cite{vaswani2017attention}. In the decoder stage, the social context is fed into the anchor decoder to predict the embeddings of $N$ anchors for each agent. 
Here the anchor embeddings can be decoded into future waypoints. At the same time, we can select the target agent feature from 
the social context, and the target agent feature is then decoded as the proposals. 
For the prediction decoder, we aim to produce $K$ proposed future 
trajectories along with their corresponding probabilities for each agent. 
Together with proposals, the anchor embeddings are fed into the prediction decoder~\cite{dong2021multi}. 
Finally, our model outputs the multi-modal trajectories and probabilities for each agent.

\subsection{Training Details}

Our loss is composed of two parts: anchor loss and prediction loss. For the anchor loss, we use 
the mean squared error and cross entropy of the future waypoints for each agent. 
For the prediction loss, we adopt the same strategy while for all future timesteps. We set the batch 
size as 12 for training. $K=6$ is used to predict the multi-modal future trajectories.

\section{Experiments}

\subsection{Dataset}

We evaluate our proposed approach on Argoverse 2. This dataset consists of 250,000 
scenarios with trajectory data for multiple object types (vehicle, pedestrian, 
motorcyclist, cyclist and bus). Each scenario is 11s long and is annotated 
at 10 frames per second. Each sample contains an agent marked as "focal track" that is needed 
to be predicted. Our trajectory prediction horizon is 60 timesteps (6s) based on the observed 
history of 15 timesteps (1.5s).


\subsection{Quantitative Results}

We follow the same evaluation protocol as defined in the challenge, and use the metrics including average displacement error (ADE), final displacement error (FDE), miss rate (MR), brier-minADE and brier-minFDE to evaluate the performance.
The result on test set is shown in Table~\ref*{table1}. For the final submission, we independently train $M$ models and obtain $MK$ candidate trajectories. At the inference stage, we perform the K-means clustering for all endpoints of the candidate trajectories. The trajectories and their probabilities in the same cluster are averaged and $K$ predictions are generated. It is worth noting that our approach performs particularly well (ranks the 1st place) on metric minADE ($K=6$), indicating that our model is stable among all modalities. 

\begin{center}
        \input{sheet1.tex}
        \end{center} 
     
Furthermore, we report our baseline performance for each object category on the validation set of Argoverse 2, as shown in Table~\ref*{table2}. 
We observe that the results of cyclist and bus are relatively inferior to other object types, which may result from the fact that they have different behaviors and insufficient training samples.
\begin{center}
        \input{sheet2.tex}
        \end{center} 

\subsection{Qualitative Results}

\begin{figure}[htbp]
\centering  
\includegraphics[width=0.45\textwidth, height=5cm]{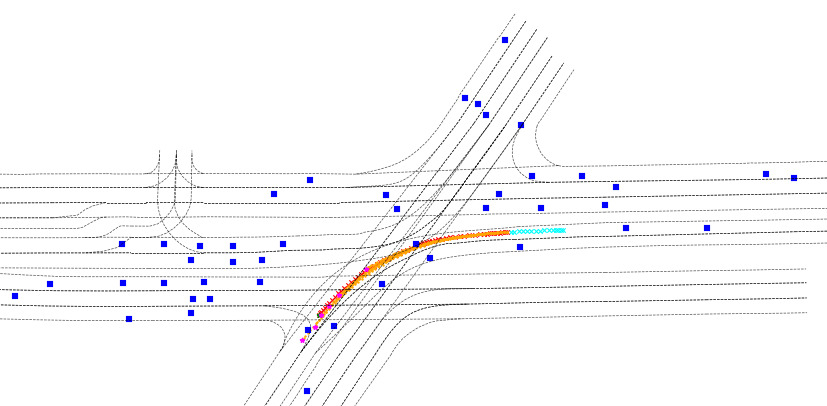}
\includegraphics[width=0.45\textwidth, height=5cm]{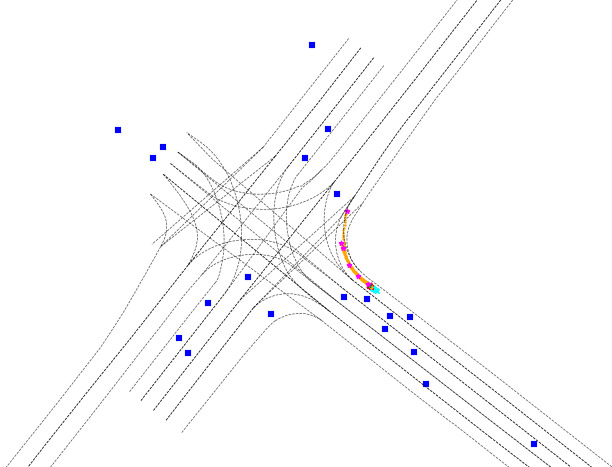}
\caption{Visualization of two representative scenes. At the intersections, agents usually tend to have diverse maneuvers. 
Our approach can provide multiple reasonable predicted trajectories.}
\label{fig2}
\end{figure}

As demonstrated in Figure~\ref{fig2}, we visualize two prediction examples of our approach on the validation set of Argoverse 2.
The black lines and blue points represent the road lanes and 
other agents around the target agent. The cyan points denote the observed 
history trajectory of the target agent, and the yellow points are the multi-modal 
outputs of our model and the red line is the ground truth future trajectory. 
The endpoints of predictions and ground truth are colored magenta and green.
As can be seen, when encountering crowded intersections, our model can 
make correct predictions no matter the target agent is moving or static.

\vspace{3cm}

\bibliographystyle{unsrt}
\bibliography{ref}

\end{document}

%% file: sheet1.tex
\begin{table}[htbp]
\centering
\caption{Evaluation on the test set of Argoverse 2.}
      \begin{threeparttable}
      {
        \begin{tabular}{c|c|c|c|c|c}
        \toprule
               & \multicolumn{1}{c|}{minADE} & \multicolumn{1}{c|}{minFDE} & \multicolumn{1}{c|}{Miss Rate} & \multicolumn{1}{c|}{brier-minADE} & \multicolumn{1}{c}{brier-minFDE} \\
        \midrule
              K = 1 & 1.8369 & 4.9779 & 0.6212 & - & - \\
        \midrule
              K = 6 & 0.6882 & 1.3850 & 0.1894 & 2.3183 & 1.9547 \\
        \bottomrule
        \end{tabular}
      }%
      \label{table1}%
      \end{threeparttable}
\end{table}%
      

%% file: sheet2.tex
\begin{table}[htbp]
\centering
\caption{Evaluation of our baseline on the validation set of Argoverse 2.}
      \begin{threeparttable}
      {
            \begin{tabular}{l|c|c|c|c}
            \toprule
                  & minADE & minFDE & brier-minADE & brier-minFDE \\ 
            \midrule
            Vehicle   & 0.8271 & 1.4478 & 1.5051       & 2.1258       \\ 
            Pedestrian   & 0.3482 & 0.6266 & 1.0196       & 1.2890       \\ 
            Cyclist   & 1.0017 & 1.7990 & 1.6721       & 2.4694       \\ 
            Motorcyclist & 0.6643 & 1.2257 & 1.3401       & 1.9015       \\ 
            Bus   & 1.0320 & 1.4625 & 1.7144       & 2.1450       \\
            \bottomrule
            \end{tabular}
            
        }%
      \label{table2}%
      \end{threeparttable}
\end{table}%